\documentclass[conference]{IEEEtran}
\IEEEoverridecommandlockouts
\IEEEpubid{\makebox[\columnwidth]{ 979-8-3503-5067-8/24/\$31.00~\copyright2024 IEEE \hfill} 
\hspace{\columnsep}\makebox[\columnwidth]{ }}

\usepackage{cite}
\usepackage{amsmath,amssymb,amsfonts}
\usepackage{algorithmic}
\usepackage{graphicx}
\usepackage{wrapfig}
\usepackage{textcomp}
\usepackage{enumerate}
\usepackage{comment}
\usepackage{tabularx}
\usepackage{booktabs}
\usepackage{url}
\usepackage{units}
\usepackage{csquotes}
\usepackage[T1]{fontenc}
\usepackage{xcolor}
\def\BibTeX{{\rm B\kern-.05em{\sc i\kern-.025em b}\kern-.08em
    T\kern-.1667em\lower.7ex\hbox{E}\kern-.125emX}}

\newcommand{\set}[1]{\ensuremath\mathcal{#1}}
\newcommand{\vect}[1]{\ensuremath\mathbf{#1}}
\newcommand{\card}{\ensuremath\vect{c}}
\newcommand{\mycards}{\ensuremath\set{C}}

\newcommand{\pref}{\succ}

\begin{document}

\title{Learning With Generalised Card Representations \\for “Magic: The Gathering”
}

 \author{\IEEEauthorblockN{Timo Bertram\IEEEauthorrefmark{1}\IEEEauthorrefmark{2}}
 \IEEEauthorblockA{\IEEEauthorrefmark{1}\textit{Dept.\ of Computer Science}\\
 \textit{Johannes-Kepler Universit\"at} \\
 \textit{Linz, Austria} \\
 \url{tbertram@faw.jku.at}}
 \and
 \IEEEauthorblockN{Johannes F\"urnkranz\IEEEauthorrefmark{1}\IEEEauthorrefmark{2}}
  \IEEEauthorblockA{\IEEEauthorrefmark{2}\textit{LIT Artificial Intelligence Lab}\\
 \textit{Johannes-Kepler Universit\"at} \\
 \textit{Linz, Austria} \\
 \url{juffi@faw.jku.at}}
 \and
 \IEEEauthorblockN{Martin M\"uller}
 \IEEEauthorblockA{\textit{Dept.\ of Computing Science}
 and \textit{Amii}\\
 \textit{University of Alberta}\\
 \textit{Edmonton, Canada} \\
 \url{mmueller@ualberta.ca}}
 }
\maketitle
\IEEEpubidadjcol

\begin{abstract}

A defining feature of collectable card games is the deck building process prior to actual gameplay, in which players form their decks according to some restrictions. Learning to build decks is difficult for players and models alike due to the large card variety and highly complex semantics, as well as requiring meaningful card and deck representations when aiming to utilise AI. In addition, regular releases of new card sets lead to unforeseeable fluctuations in the available card pool, thus affecting possible deck configurations and requiring continuous updates. Previous Game AI approaches to building decks have often been limited to fixed sets of possible cards, which greatly limits their utility in practice. In this work, we explore possible card representations that generalise to unseen cards, thus greatly extending the real-world utility of AI-based deck building for the game “Magic: The Gathering”. We study such representations based on numerical, nominal, and text-based features of cards, card images, and meta information about card usage from third-party services. Our results show that while the particular choice of generalised input representation has little effect on learning to predict human card selections among known cards, the performance on new, unseen cards can be greatly improved. Our generalised model is able to predict 55\% of human choices on completely unseen cards, thus showing a deep understanding of card quality and strategy.
\end{abstract}


\section{Introduction}

Modern board and card games are popular domains for AI research due to their inherent structure and wide popularity. While super-human play has been achieved in some games such as Go, chess, DOTA and poker \cite{silver2017mastering, berner2019dota, brown2019superhuman}, many others still provide numerous obstacles. Providing additional difficulty, commercial games, and especially collectable card games such as \textit{Hearthstone} \cite{dockhorn2019introducing} and \textit{Magic: The Gathering (MTG)}, are constantly evolving due to card changes and the release of novel content. This implies that any model for such games must be able to continuously adapt to new releases to maintain usefulness. In this work, we focus on one crucial aspect of the popular card game \textit{Magic: The Gathering}; deck building. Before card play begins, players build their deck of cards by selecting from a large pool of available choices. This initial stage of the game is crucial for success, as good performance is not achievable without strong and general deck building strategies. We focus on learning a generalised model for \textbf{drafting}, a sub-area of deck building, in \textit{Magic: The Gathering} that is readily expandable to new content and can accurately predict human decisions. To achieve this, we incorporate the \textbf{contextual preference ranking} (CPR) framework of Bertram et al. \cite{bertram2021predicting} and adapt it in several ways:

\begin{itemize}
    \item The original work on CPR is limited to a single, fixed set of cards, using a one-hot encoding to represent inputs to a Siamese neural network. We explore different generalised card representations for use in CPR, thus extending the method to arbitrary and potentially previously unseen cards.
    \item We train a large model for drafting in \textit{Magic: The Gathering} on a large, heterogeneous dataset of 100 million decisions. The model obtains general knowledge of card semantics, showing usefulness for drafting decks with new cards.
    \item We explore fine-tuning the model to quickly adapt to newly released card sets, achieving rapid performance improvements.
\end{itemize}

\section{Related Work}

We build on the \textbf{contextual preference ranking} (CPR) approach \cite{bertram2021predicting} by making it more general and widely usable, while also improving upon its overall accuracy. CPR functions by constructing preferences between choices within an explicitly stated context in which the decision was made. Data on such decisions-in-context can be coded as triplets, which are used to train \textbf{Siamese neural networks} (SNNs) \cite{bromley1993signature}. As one such case, card drafting decisions (see Section~\ref{sec:drafting}) in \textit{Magic: The Gathering} can be regarded as preferences among the different available cards in the context of all previously chosen one. In the initial CPR research \cite{bertram2021predicting}, this approach was shown to outperform classical fully connected neural networks \cite{ward2021ai}.

Other research on drafting, or deck building in general, is largely based on reinforcement learning or genetic algorithms \cite{bjorke2017deckbuilding}. For the game of \textit{Legends of Code and Magic} (LOCM) \cite{kowalski2023summarizing}, a simple card game developed for testing AI agents, earlier work \cite{miernik2021evolving,vieira2020drafting} separates the drafting and playing phases and learns to draft while approximating the strength of the resulting decks through heuristic-based gameplay. Such approaches depend strongly on the chosen gameplay strategies and therefore limit the depth of strategy. Later, end-to-end approaches for drafting and playing were developed \cite{xi2023mastering}, which led to drafting strategies decoupled from heuristics. However, in the LOCM test-bed, card complexity is low and gameplay is simple, which limits applicability to real games. Xiao et al. \cite{xiao2023mastering} used a similar algorithm, an end-to-end reinforcement learning agent combining card selection and gameplay, for the highly successful commercial game of \textit{Hearthstone}. This approach yields impressive results and appears to be a promising direction when fast-forward game simulations are possible. However, in order to allow exploration of the enormous game space, the card pool and selection options were strongly restricted, thus limiting overall usefulness. So far, it appears that end-to-end reinforcement learning of drafting and gameplay is not yet practical for unrestricted game environments when regarding ever-changing domains. Our work combats this problem from a different angle: instead of attempting to learn extremely complex relations with no prior knowledge, we rely on human behaviour and model decisions seen in historic game data in a generalised manner.

Overall, human players of \textit{Magic: The Gathering} have shown significant interest in using drafting agents as tools to aid their decision-making. Popular tools for this task\footnote{\url{https://draftsim.com/arenatutor/}, \url{https://mtga.untapped.gg/companion}, and \url{https://mtgaassistant.net/}} attempt to solve the problem though hand-constructed rules, and thus rely on humans in the design and prevent the usage of AI models.
One of these employs a classical fully connected neural network trained with supervised learning \cite{ward2021ai}, but received worse predictive accuracy than CPR \cite{bertram2021predicting}. Peer-reviewed research results on the other tools, or on independent projects\footnote{\url{https://github.com/RyanSaxe/mtg}}, are not available at this time. Finally, general work on card games experimented with different card representations \cite{zilio2018neural, pawlicki2014prediction}, opting to represent features or images of cards for classification and regression, which shows promising results.

\section{Drafting in collectable card games}
\label{sec:drafting}
Most collectable card games (CCG) provide two overarching modes of gameplay: \textit{Constructed} and \textit{Limited}.
This includes commercial games such as \textit{Hearthstone}, \textit{Magic: The Gathering}, \textit{Lorcana} or \textit{Flesh and Blood}, as well as simple AI counterparts such as \textit{Legends of Code and Magic}. To simplify the following explanation, we focus on \textit{Magic: The Gathering (MTG)}, but most rules apply to other games with minimal adaptions.

In \textbf{constructed} modes, each player independently builds a deck of at least 60 cards in advance of playing the game. Players possess mostly free rein over which cards to include in their deck, leading to an enormous space of potential configurations. Theoretically, the possible deck space is infinitely large, as no upper bound on legal card quantity exists. In practice, the vast majority of players select the smallest legal number of 60 cards to maximise the consistency of their decks by minimising the variance of drawn cards. At the time of writing, the number of unique cards to choose from ranges from 3,037 cards in the smallest version of the game to 26,524 in the largest format. Based on this, and disregarding some specifics of deck building, the number of unique decks players could potentially build lies between ${3037 \choose 60} >10^{126}$ and ${26524 \choose 60 }>10^{183}$. While an enormous space, the majority of such deck compositions are impractical and high-level human competition mostly converges to fewer than 100 deck archetypes with slight individual adaptions. How one could train models that emulate such an enormous reduction in deck space remains an open research question.

\textbf{Limited} games, or in this context \textbf{drafts}, incorporate deck building into the actual game. Before play commences, players take turns to build their decks sequentially. Players start the draft with a choice of 15 initial cards and select a single one, passing all remaining cards to the other players until all initial cards are selected. This process repeats two more times with 15 cards per player, thus leading to a total of 45 cards and 42 choices with 15 to 2 options each for every participant. Therefore, within one draft, one player's decisions result in $(15!)^3>10^{36}$ different possible final deck configurations. While still large, the sequential nature of decisions leads to smaller, more manageable, and largely independent tasks. In this work, we focus on such drafting environments, viewing each decision as independent and disregarding the effect of card choices on other players. As such, we solely focus on each individual choice in the context of the player's previous decisions, enabling the use of \textbf{CPR}.

\begin{figure*}
    \centering
    \includegraphics[width = \textwidth]{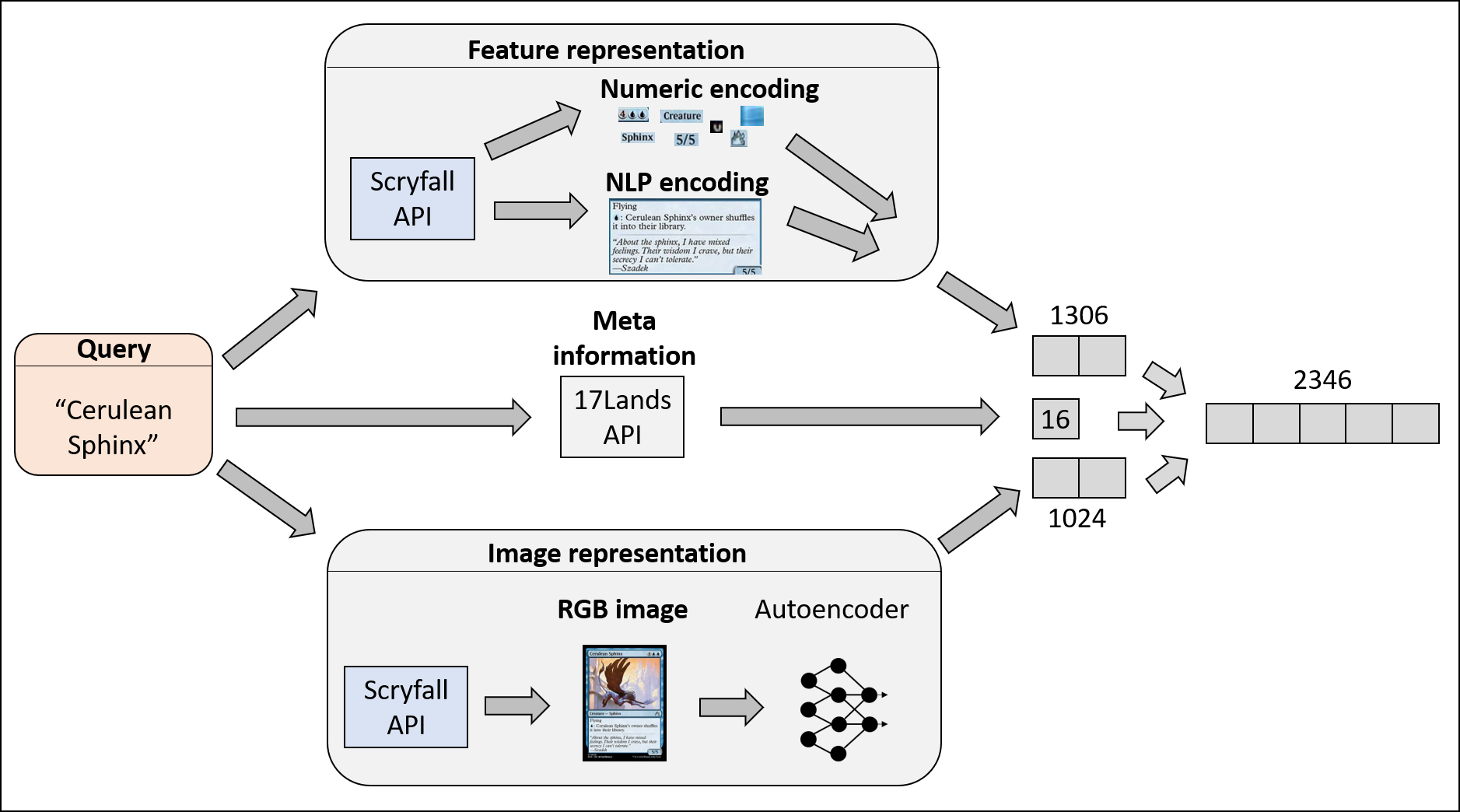}
    \caption{High-level overview of generating card representations according to Section~\ref{sec:Input representation}. The representation-variant shown here is \textit{Features + Meta + Image}, simpler representations only use parts of the pipeline.}
    \label{fig:overview_representation}
\end{figure*}
\section{Adapting Contextual Preference Ranking}
\label{sec:Adapt}

\subsection{Contextual Preference Ranking}
\label{sec:CPR}

\textbf{Contextual preference ranking} (CPR) is a framework for learning to predict the better of two choices $\card_j$ and $\card_k$ in the context of a set of previously made decisions $\mycards$, formally denoted as 
\begin{equation}
        \left(\card_j \pref \card_k \mid \mycards\right).
\end{equation}
In our setting, the context $\mycards$ represents the (incomplete) deck of cards a player is holding, $\card_j$ and $\card_k$ are two possible additions to it, while the preference encodes which one of the two is a better fit to $\mycards$.

Bertram et al. \cite{bertram2021predicting} tackled this problem by training a Siamese neural network (SNN) from a database of human card selection decisions.
The key idea is to train a single neural network $N(.)$ that maps card sets and individual cards into a uniform embedding space by minimising the triplet loss 
\begin{equation}
\label{eq:triplet}
    L_{\textrm{triplet}}(\vect{a},\vect{p},\vect{n}) = \max\left(d(\vect{a},\vect{p})-d(\vect{a},\vect{n}) + m,0\right)
\end{equation}
over three weight-sharing copies of the network, where the anchor $\vect{a} = N(\mycards)$ represents the embedding of the context $\mycards$, $\vect{p} = N(\card_j)$ the embedding of the preferred choice $\card_j$, and $\vect{n} = N(\card_k)$ the other choice. The parameter $m$ represents a desired margin (typically set to $m=1$), and $d$ is a distance metric (typically the Euclidean distance).

In testing, the model embeds the context $\mycards$ and all possible choices $\card_i$, and ranks choices by their embedded distance to $\mycards$. A card
\begin{equation}
\label{eq:card-selection}
    \card^* = \arg \min_{\card_i} d(N(\card_i),N(\mycards))
\end{equation} 
with minimal distance is chosen as the best addition to the current deck $\mycards$.

We follow this framework but with a generalised implementation. In the former work, the 265 unique cards were represented with a one-hot encoding for input into the SNNs. This input representation can not be adapted for prediction on unseen cards, and separate models would need to be trained for each new set of cards. The majority of our work focuses on this problem of training on arbitrary sets and generalising to unseen inputs.

\subsection{Input representation}
\label{sec:Input representation}

In \textit{MTG}, cards are released in so-called \textit{expansions} or \textit{sets}, which feature 200--300 unique cards. The majority of these cards are completely new and often feature new rules and mechanics. In order to generalise CPR for this type of application, we study different representations of cards and explore their advantages and disadvantages.

\subsubsection{Random vectors}
Representing cards by randomly generated vectors of arbitrary size enables an infinite space of possible inputs. However, this representation prevents generalising knowledge to unseen cards and learning semantics and is thus only useful for in-sample cards. Still, we experiment with random vectors as card representations to provide insight into the importance of representation choice for seen sets. In these experiments, each unique card is encoded as a randomly generated vector of size 1024.

\subsubsection{Hand-coded features}

\begin{figure}[h]
\centering
    \includegraphics[width = \columnwidth]{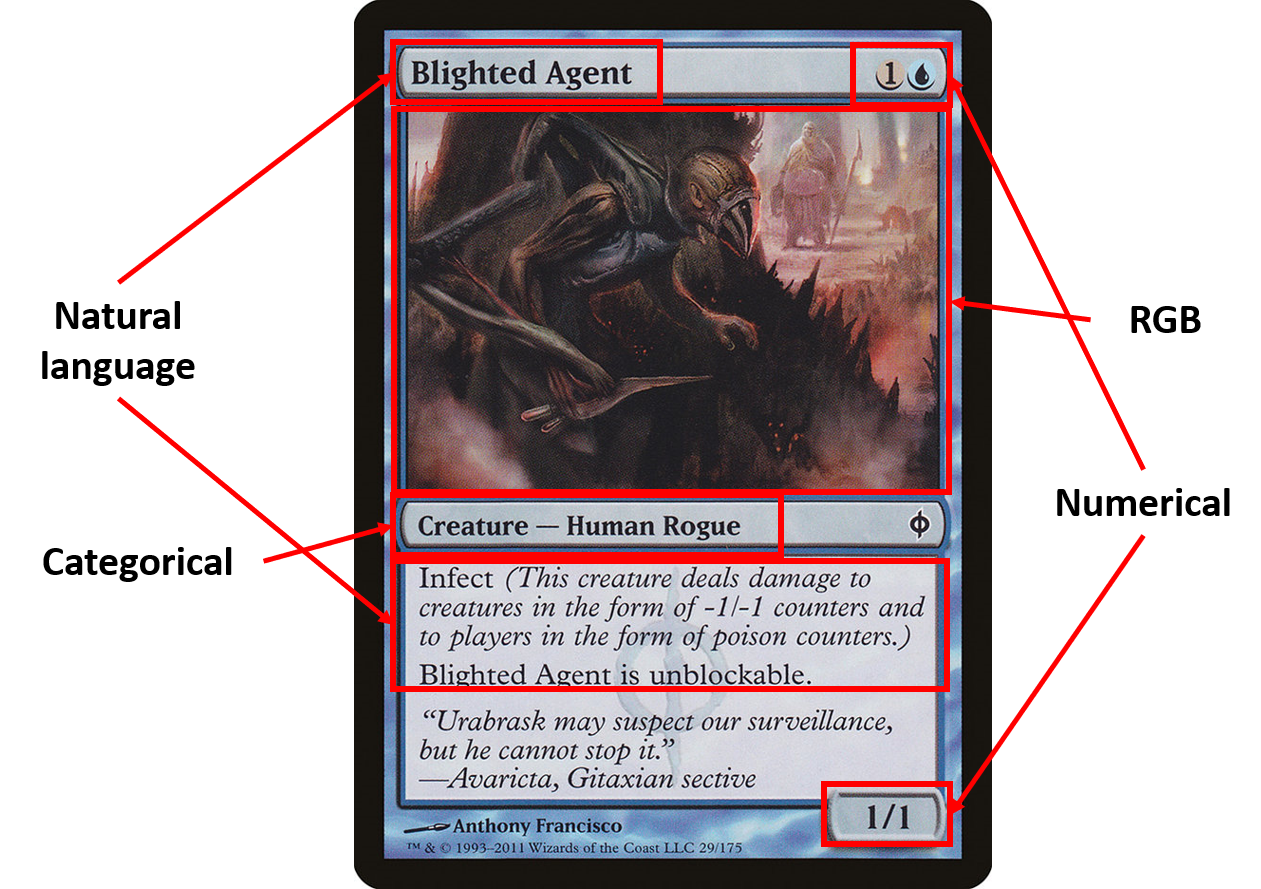}
    \caption{Anatomy of a \textit{Magic: The Gathering} card. Cards consist of a number of different features of varying importance and representation.}
    \label{fig:magic card}
\end{figure}

To generalise across cards and sets, we can encode cards by their unique \textit{features}, which capture the rules and semantics associated with them. \textit{Magic} cards possess numerous features including numerical values, categorical features of large categories, and a card text of great importance (Figure~\ref{fig:magic card}). To a lesser extent, they contain an RBG image, which will be regarded in the following section. For the \textit{Features} representation, we encode a card as a vector of all numerical and categorical features and append a text-embedding of the full card text generated with a sentence transformer \cite{reimers2019sentence}.

\subsubsection{Image Representation}

\begin{figure}
    \centering
    \includegraphics[width = \columnwidth]{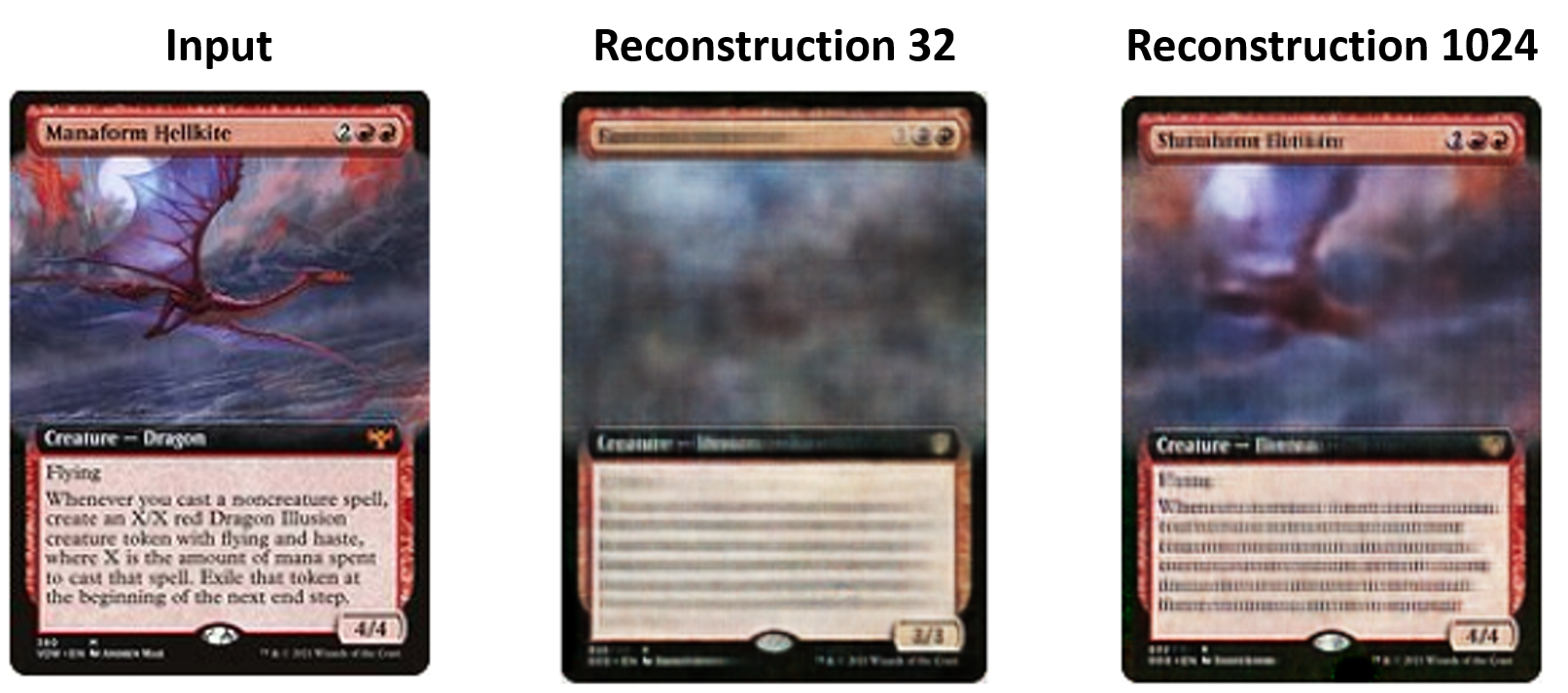}
    \caption{Inputs and reconstructions of trained autoencoders. Input and outputs are of shape $3\times 936\times 672$, the latent spaces are 32-dimensional and 1024-dimensional respectively. The latent card representations are used as compressed input to the SNN (see Figure~\ref{tab:accuracy inputs})}.
    \label{fig:autoencoder}
\end{figure}

When simply encoding a card by its pixel-wise RGB values, no handcrafted features are required. However, this tasks the model to learn the complex semantics of a card purely from visual cognition. Text recognition from images is challenging \cite{wang2012end} and likely leads to a loss of information compared to an explicit encoding. In addition, full-scale RGB images of cards are large ($3\times 936\times 672)$, leading to an explosion in input size. To combat this, we train a basic convolutional autoencoder \cite{masci2011stacked} to construct latent representations of card images, which can be used as compressed representations of inputs to the SNN. In our experiments, we found strong influence, both visually (see Figure~\ref{fig:autoencoder}) and numerically (see Table~\ref{tab:accuracy inputs}), of the bottleneck dimensionality on the received reconstructions. We present the results with latent space dimensionalities of 32 and 1024, finally choosing 1024 dimensions for later experiments to minimise information loss.

\subsubsection{Statistical Meta Information on Cards}
Meta information about card use by human players, such as pick-rate and player win-rates when using the card, is publicly available\footnote{\url{https://www.17lands.com/card_data}}. Such usage information provides direct insight into human decisions for the cards involved. This is an especially useful feature when little useful context is available, e.g. when comparing similar cards, or at stages of deck building when few cards have been selected. We provide an experiment where we train a model solely on the meta-information (Table~\ref{tab:accuracy inputs}), but generally, such statistics should be viewed as additional information rather than a stand-alone encoding. Noteworthy, card statistics are only available when a set has seen sufficient server play and is thus impossible to use for entirely new sets. In our experiments, we encode statistics as a scalar vector of size 16.

\subsection{Encoding networks}
\label{sec:encoding_networks}

Regardless of the specific encoding used, adapting CPR from one-hot to generalised encodings creatures the additional challenge of deriving a representation for card decks from the representation of single cards \cite{bertram2023weighting}. With one-hot encodings of cards, a deck can simply be represented as the sum of card encodings without loss of information. However, with feature-based card encodings, a sum-based encoding of a deck makes it impossible to reconstruct individual cards. To prevent this, we represent a deck of cards as a two-dimensional array of size $45 \times n$, where $45$ is the maximum number of cards in the deck and $n$ is the representation size of individual cards. Slots for not-yet-picked cards are set to zero. When decks and cards are used as inputs to the SNN, this first requires translating them to a common representation (see Figure~\ref{fig:overview}), which is achieved by adding two separate encoding networks that output 512-dimensional representations, of cards and decks respectively, to the training pipeline. These encoding networks are trained end-to-end with the SNN.

\section{Experimental Setup}
\label{sec:Experiments}

\subsection{Data}
We evaluate the influence of the outlined adaptions to CPR on the accuracy of the resulting model. In addition to the changes mentioned in Section~\ref{sec:Adapt}, the experiments feature a larger variety of card sets from different sources, collectively encoding approximately 100 million total card picks. Table~\ref{tab:sets} shows all used datasets along with their respective release dates and training set sizes. We used all released datasets up to and including \textit{LTR}. Data of M19 is obtained from \textit{DraftSim}\footnote{\url{https://draftsim.com/draft-data/}}, all other datasets are provided by \textit{17Lands}\footnote{\url{https://www.17lands.com/public_datasets}}. Each dataset was split into 80\% training and 20\% test set. Note, that these numbers show the individual card picks. In our preference-based formulation, each pick results in 1--14 preferences, thus increasing the total number of triplets by almost an order of magnitude.

\begin{table}[h]
\caption{All card sets used with their respective release data and training set size.
}
\label{tab:sets}
\centering
\begin{tabular}{lcr}
\toprule
Set & Release date & Training size\\
\toprule
LTR & 23-06-2023 & 684,724\\
MOM & 21-04-2023 & 5,085,312\\
SIR& 21-03-2023 & 2,422,668\\
ONE & 10-02-2023& 5,260,169\\
BRO & 18-11-2022 & 4,153,162\\
DMU& 09-09-2022& 7,887,976 \\
HBG& 07-07-2022& 1,680,866\\
SNC& 29-04-2022& 5,753,840\\
NEO& 18-02-2022& 5,122,921\\
VOW& 19-21-2021& 4,012,657\\
MID& 24-09-2021& 3,363,477\\
AFR& 23-07-2021& 959,794\\
STX& 23-04-2021& 3,809,102\\
M19& 07-07-2018& 29,094,192\\
\bottomrule
\end{tabular}
\end{table}

\subsection{Evaluation Measure}
We use the \emph{top-1 accuracy} when ranking all possible picks to evaluate the models. To create the ranking, we embed the current deck and all possible options, sort all options by their distance to the deck in the embedding space, and choose the card with the minimal distance as shown in~\eqref{eq:card-selection}. The selected card is compared to the human choice captured in the dataset, and the percentage of matching choices is reported. Note that, as this aims to predict human choices, it is generally not possible to achieve perfect accuracy, as human decisions are not necessarily correct or consistent. 

\subsection{Siamese Neural Network configuration}

\begin{figure}
    \centering
    \includegraphics[width = \columnwidth]{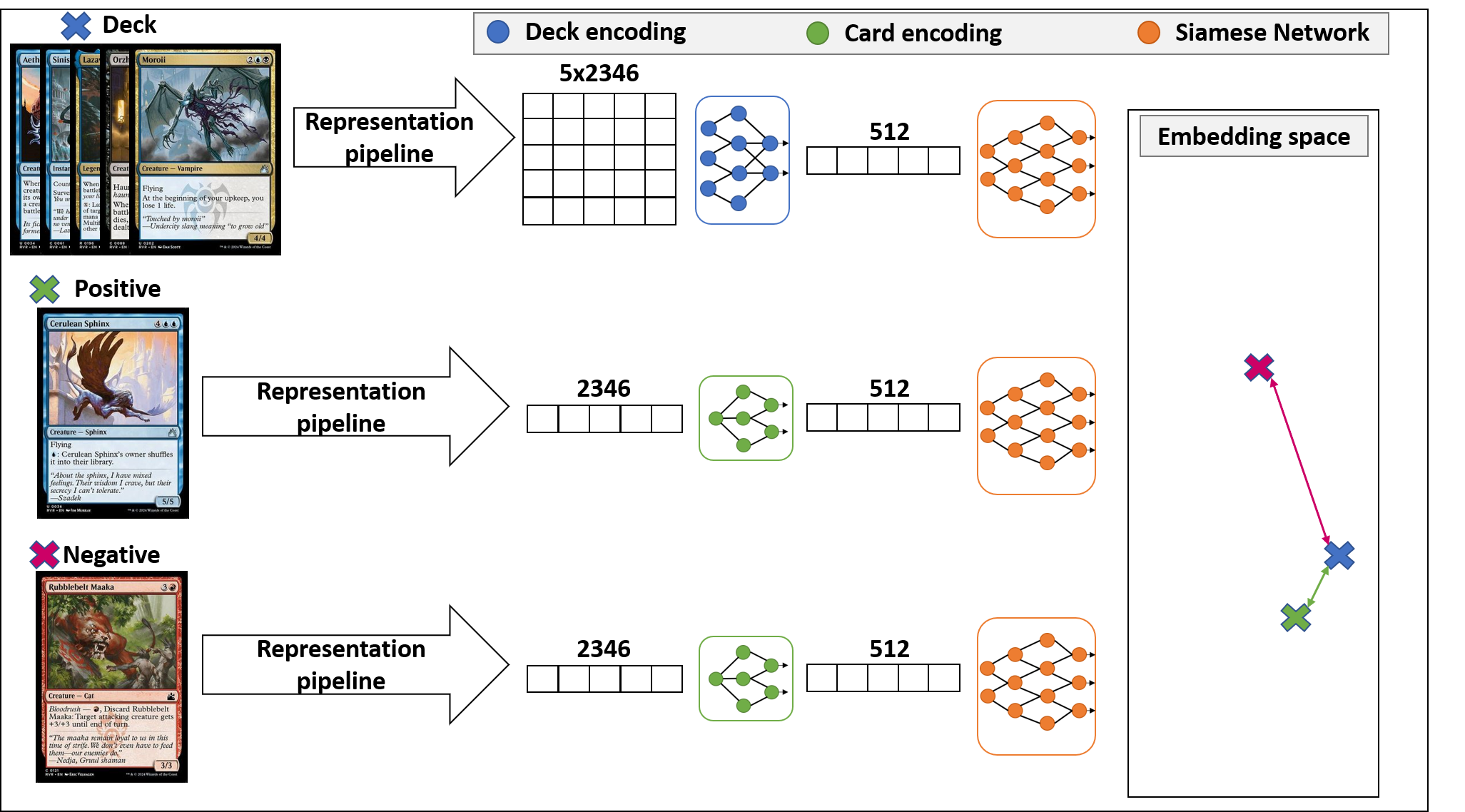}
    \caption{General overview of training. Refer to Figure~\ref{fig:overview_representation} for the explanation of the representation pipeline. Training loss is computed based on Equation~\eqref{eq:triplet}.}
    \label{fig:overview}
\end{figure}

As explained in Section~\ref{sec:encoding_networks}, using CPR with generalised card representations requires separate encoding networks for cards and decks. The card encoding network uses 4 fully connected layers with 1024 neurons, connected by dropout, normalisations and ELU activations. The deck encoding network consists of 6 convolutional layers with 1 to 16 filters, normalisations, max-pooling and ELU activations with a single fully connected output layer. Both encoding networks feed into the same Siamese neural network, which uses 5 fully connected layers with 512 neurons, normalisations, dropout and ELU activations. The final layer possesses 512 outputs and uses the tangens hyperbolicus activation, thus creating a 512-dimensional embedding space of range $[-1,1]$. For a general overview, refer to Figure~\ref{fig:overview}.

\section{Results}
Experiments in this section are split in two: Section~\ref{sec:representation} investigates the different input encodings (see Section~\ref{sec:Input representation}) and Section~\ref{sec:general-models} evaluates fine-tuning a pre-trained model on previously unseen cards, i.e. simulating the release of a new card set.

\subsection{Representation}
\label{sec:representation}

To investigate the influence of card representation (see Section~\ref{sec:Input representation}) on the received models, we conduct a series of experiments with SNNs that only differ in input representation. To reduce the computational effort of this experiment, models are trained solely on the \textit{NEO} set. The accuracy of the received models is reported on held-out samples of the \textit{NEO} set and the average accuracy across all unseen sets. Section~\ref{sec:general-models} uses the whole dataset to maximise predictive performance and simulate real-world use.

\begin{table}[h]
\label{tab:accuracy inputs}
\caption{Top-1 testing accuracy of models with differently encoded inputs. All models are trained solely on \textit{NEO} and tested on a held-out \textit{NEO} test-set and unseen cards. Accuracy is averaged across all unseen sets.}
\resizebox{\columnwidth}{!}{
\begin{tabular}{ccccc}
\toprule
Model & Input size & \textit{NEO} test & unseen sets\\
\toprule
One-hot & 302 & 67.80\% & NaN \\
Random & 1024 &67.87\% & 23.79\%\\
Image \textit{32}& 32 & 65.93\% & 28.69\%\\
Image \textit{1024}& 1024 &68.09\% & 31.10\%\\
Meta & 16 &64.73\% & 42.14\%\\
Features & 1306 &67.76\% & 33.57\%\\
Features + Meta& 1322& 68.07\% & 34.74\% \\
Features + Image \textit{1024}& 2330& 67.81\% & 35.59\% \\
Features + Meta + Image \textit{1024} & 2346& 68.00\% & 42.87\%\\
\bottomrule
\end{tabular}}
\end{table}

In Table~\ref{tab:accuracy inputs} we see the test accuracies obtained on \textit{NEO} and the unseen sets. Interestingly, the top-1 accuracy on the seen card set is consistent across most models, apart from the two with small input spaces and thus limited information. However, accuracy on unseen sets, i.e. the ability of the model to learn general characteristics of cards, differs drastically. The \textit{Random} model is still able to accurately model decisions in \textit{NEO}, thus showing that with sufficient data on a set, generalised knowledge is not required. However, this model provides no insight into unseen cards where we receive the accuracy of random predictions\footnote{Decisions follow a uniform random distribution with 1 to 15 choices, thus random predictions would yield an accuracy of $\nicefrac{1}{15} \cdot \sum_{i=1}^{15} \nicefrac{1}{i} \approx 0.22$. The observed accuracy for the \textit{random} representation is slightly higher because some datasets are missing some picks.}. 

All other representations are able to improve upon random predictions on unseen cards. Regarding the autoencoded image representations, we receive worse accuracy with the small latent space, likely due to the generally too-small input space.  The more information-rich compression reaches high accuracy on the \textit{NEO} test-set, as well as achieving some generalisation to the unseen sets. Surprisingly, hand-engineered features only receive slightly better general results, thus leading us to speculate that with limited data, models do not generate rich semantics of cards but rather use shallow features like their colour, which are easily recognisable from the image representations. 

The \textit{Meta} model with only 16 features based on meta attributes achieves among the best accuracy on unseen sets, underlining that when limited information is available, simply using card statistics leads to reasonable results. However, combining features and meta-information is not able to utilise this, barely improving upon the \textit{Features} model. Similarly, combining \textit{Features} and \textit{Image 1024} only leads to slightly better performance on unseen cards. The reason for this is not clear to us and requires further investigation. The overall best-performing model is \textit{Features + Meta + Image}, as it is able to predict almost 43\% of decisions on completely unseen cards while retaining accuracy on \textit{NEO}.

From this experiment, the initial card representation appears largely irrelevant when CPR is used to construct an embedding space for known cards. Even random vectors as card representations and one-hot encodings yield similar results as encoding a card's semantics. However, input representation clearly matters when aiming to generalise to new, previously unseen cards. Additionally, meta-information about a card's usage, if available, is highly relevant for unknown cards and already leads to reasonable results.

\begin{figure}
    \centering
    \includegraphics[width=\columnwidth]{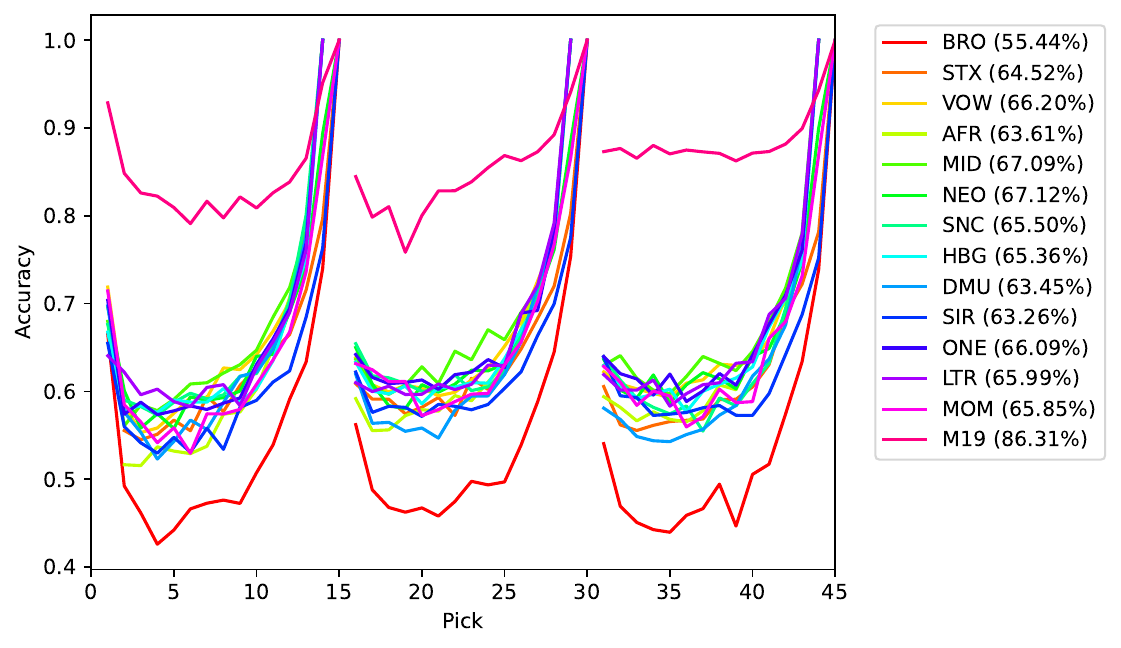}
    \caption{Accuracy of \textit{Features+Meta+Image} model trained on all sets apart from \textit{BRO}, split by set and pick. Numbers in brackets show the average accuracy per set. Graphs are separated by the three packs.}
    \label{fig:all_accuracy}
\end{figure}

\subsection{Transfer}
\label{sec:general-models}

Following these experiments, the \textit{Features+Meta+Image} model with the highest combined accuracy is chosen for further exploration. Our goal in this section is to model a real-world setting where a model is trained on all available card sets, and then tested on or adapted to a newly issued set of cards. To test this ability of transferring obtained knowledge to new card sets, we use all but one of the datasets listed in Table~\ref{tab:sets} for training. On one hand, we can expect to improve the overall performance of the model by increasing training set size and card heterogeneity, and on the other hand, this experiment is also meant to simulate the intended application scenario. To simulate a new card set being released, we hold out a single set of cards (\textit{BRO}), reporting accuracy on held-out test sets by averaging across all seen sets and explicitly reporting the test set performance on \textit{NEO} and \textit{BRO}. 

As expected, by simply increasing the sample quantity (75 million instead of 5 million) and unique card variety (2,990 cards instead of 302), the general accuracy of the model increases considerably (see Table~\ref{tab:accuracy_foundation}). Although performance on the seen sets is comparable to the previous results, the model \textit{Pre-training}, which is the same model as \textit{Features+Meta+Image\hphantom{-}1024} but trained on more data, achieves 55.44\% accuracy on the unknown \textit{BRO} set. Splitting the received accuracy over all sets and picks (Figure~\ref{fig:all_accuracy}), we see similar patterns on all sets, regardless of whether the model was trained on it or not. The accuracy of picks at the start and end of each pack is higher, while picks in the middle appear more difficult. Compared to previous work with CPR \cite{bertram2021predicting}, the accuracy on \textit{M19} is improved by more than 2 percentage points.

Building on this, we explore fine-tuning or transfer learning \cite{zhuang2020comprehensive}, with the model on two individual sets, \textit{NEO} and \textit{BRO}. When fine-tuning, the large corpus of training data is replaced with a single set of cards, thus aiming to use previously obtained general knowledge for a smaller task. For our experiment, a model is initialised with the parameters obtained from pre-training on the large corpus and trained exclusively on \textit{NEO} and \textit{BRO} without further adjustments to the training process, omitting freezing layers or other attempts to retain knowledge.

The two fine-tuning settings serve different purposes. When fine-tuning on \textit{NEO}, no new data or cards are seen. Rather, the training data is reduced to decrease the task complexity, thus gaining the opportunity to improve the accuracy. Using \textit{BRO} to fine-tune replaces previous data with unseen cards, thus investigating whether the model can quickly adjust to new information, simulating the release of a new set of cards in real applications.

\begin{figure*}
\centering
\begin{tabular}{ccc}
\includegraphics[width = 5cm, height = 4cm]{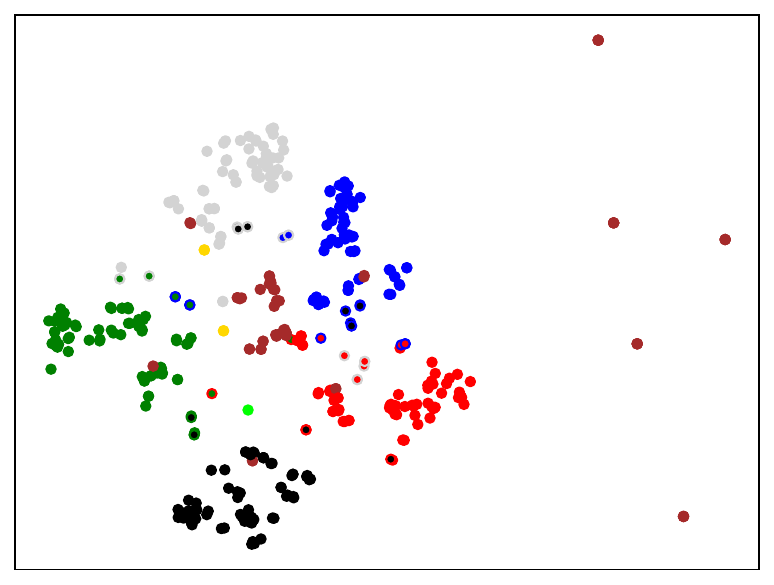} & \includegraphics[width = 5cm, height = 4cm]{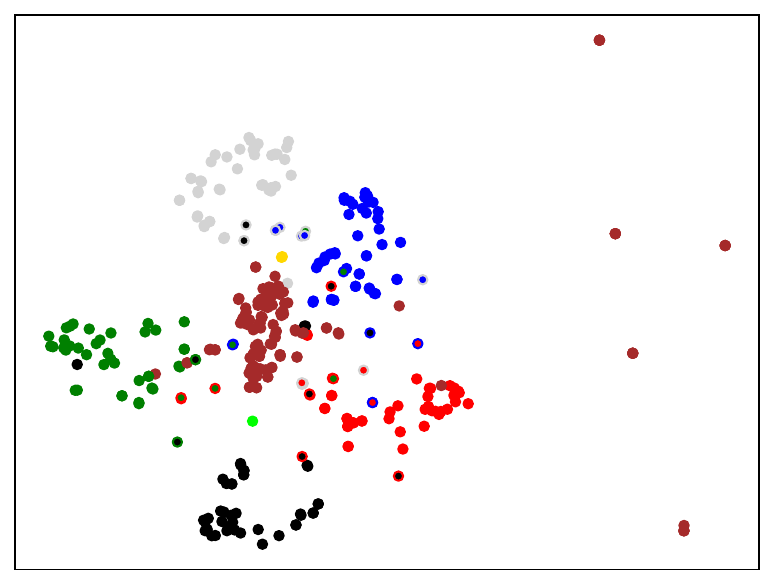} & \includegraphics[width = 5cm, height = 4cm]{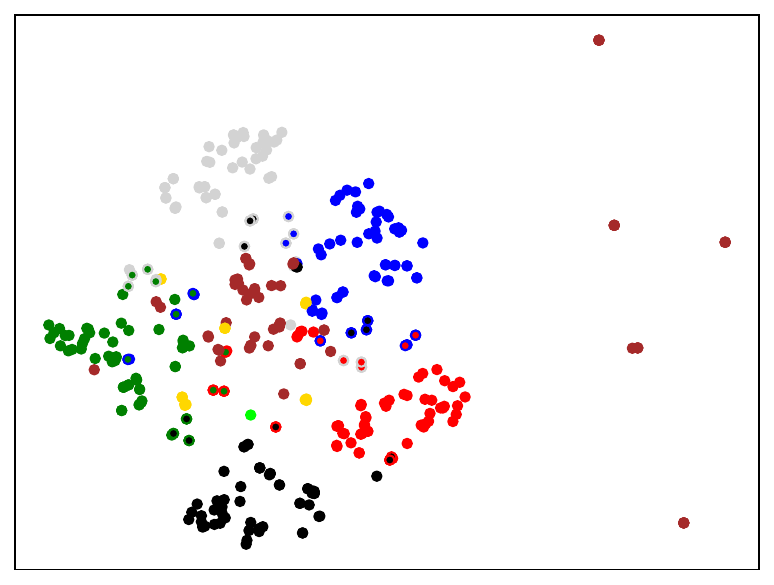}\\
(a) AFR & (b) BRO & (c) DMU\\[6pt]
\includegraphics[width = 5cm, height = 4cm]{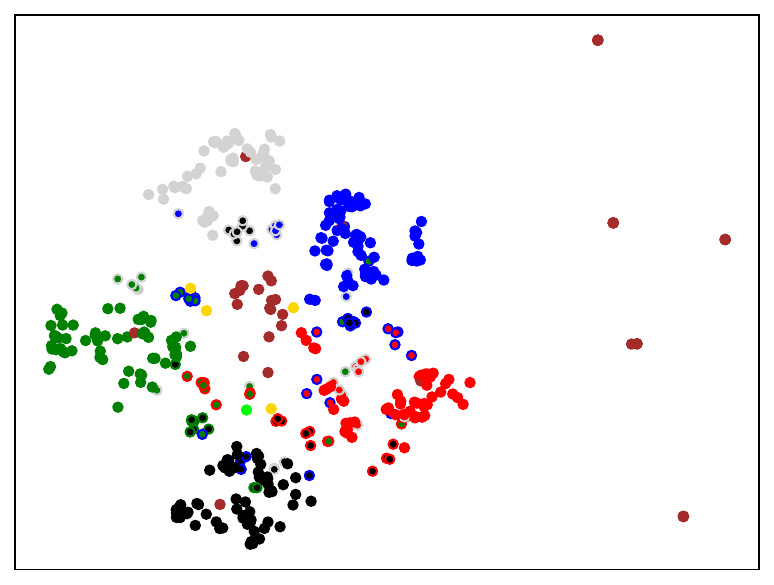} & \includegraphics[width = 5cm, height = 4cm]{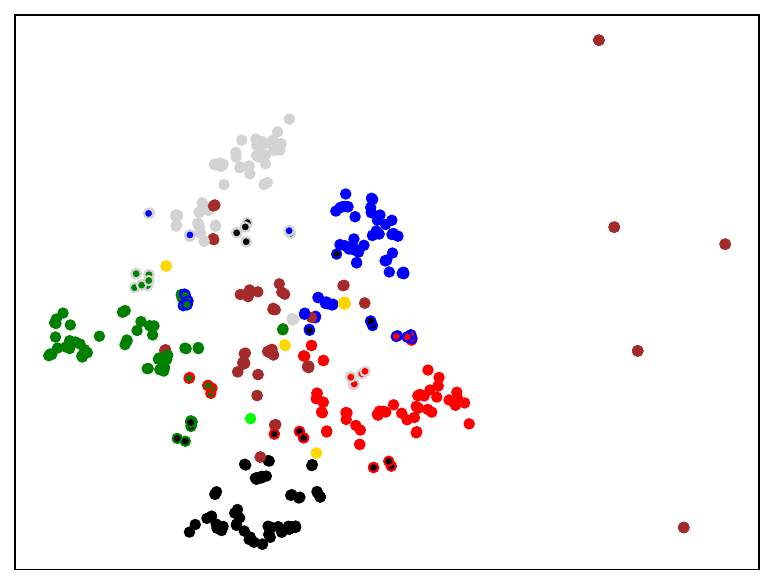}& \includegraphics[width = 5cm, height = 4cm]{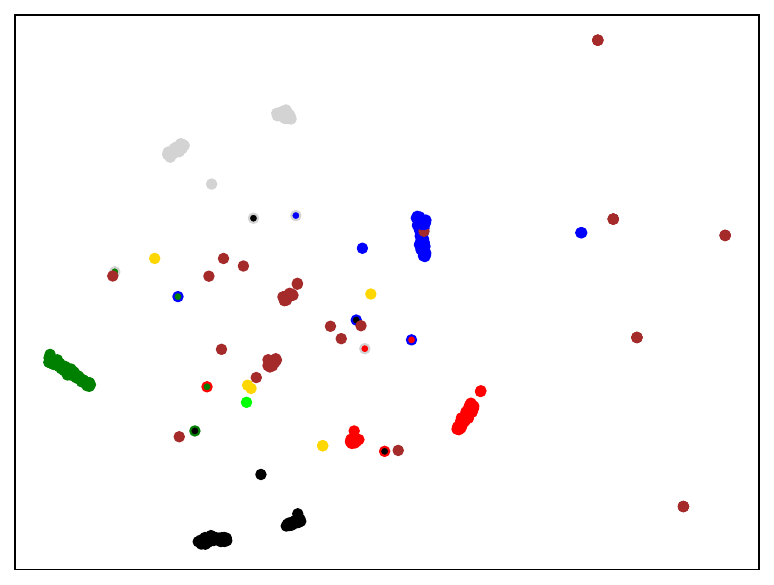}\\
(d) HBG & (e) LTR & (f) M19\\[6pt]
\includegraphics[width = 5cm, height = 4cm]{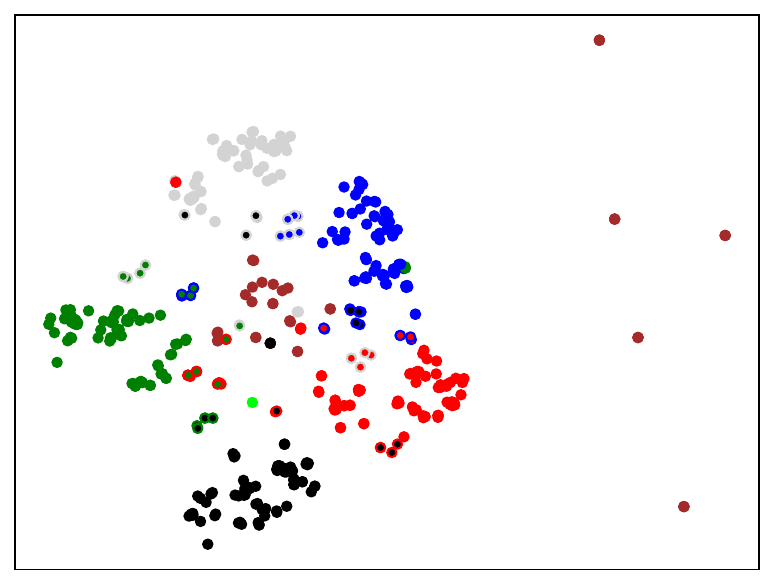} & \includegraphics[width = 5cm, height = 4cm]{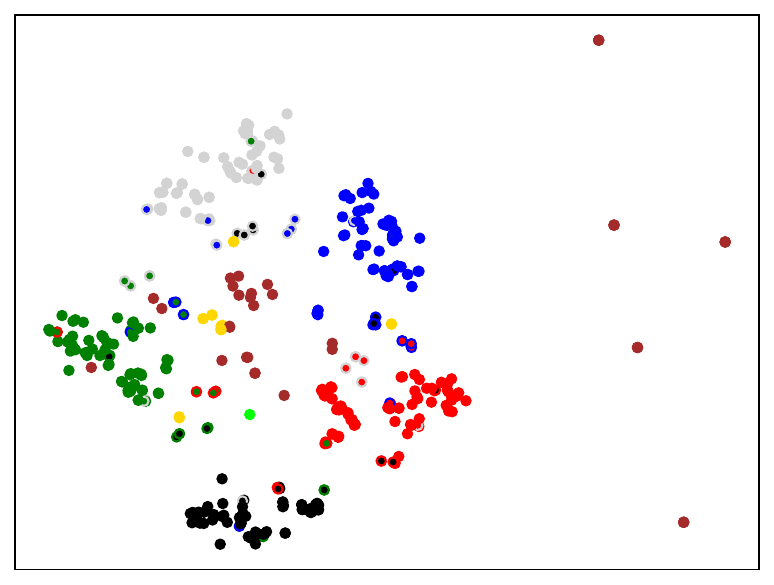} & \includegraphics[width = 5cm, height = 4cm]{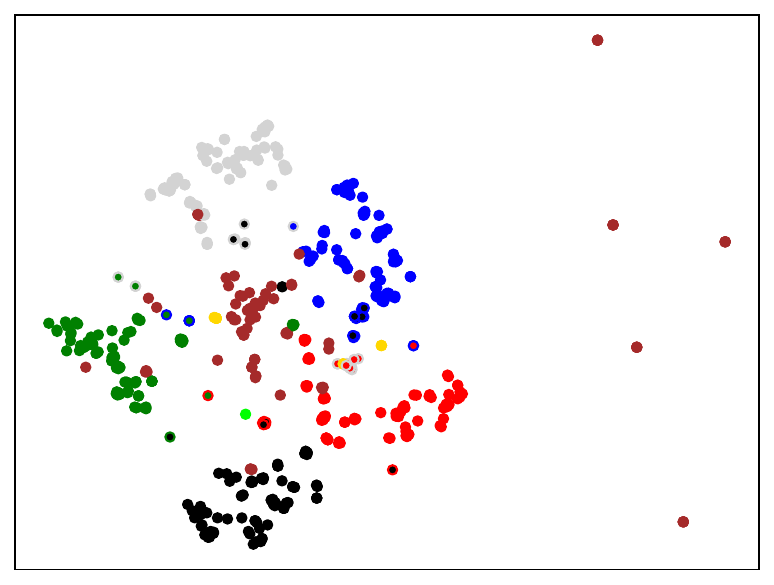}\\
(g) MID & (h) MOM & (i) NEO\\[6pt]
\includegraphics[width = 5cm, height = 4cm]{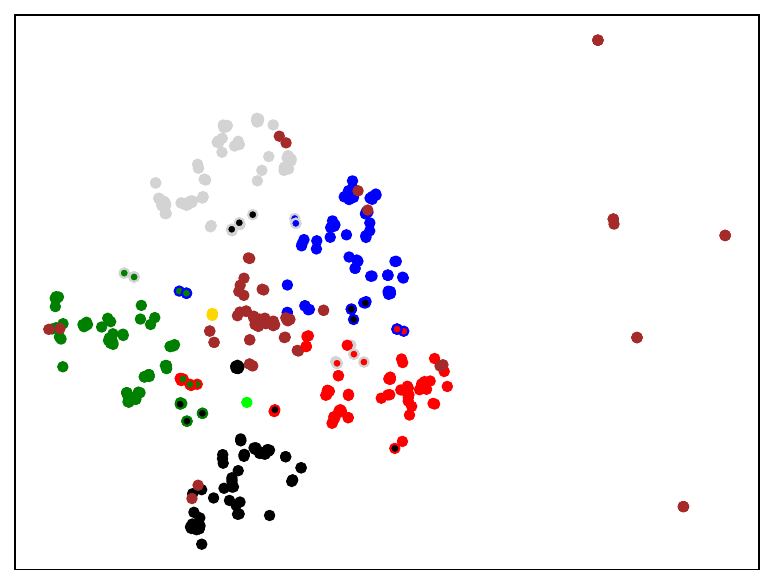} & \includegraphics[width = 5cm, height = 4cm]{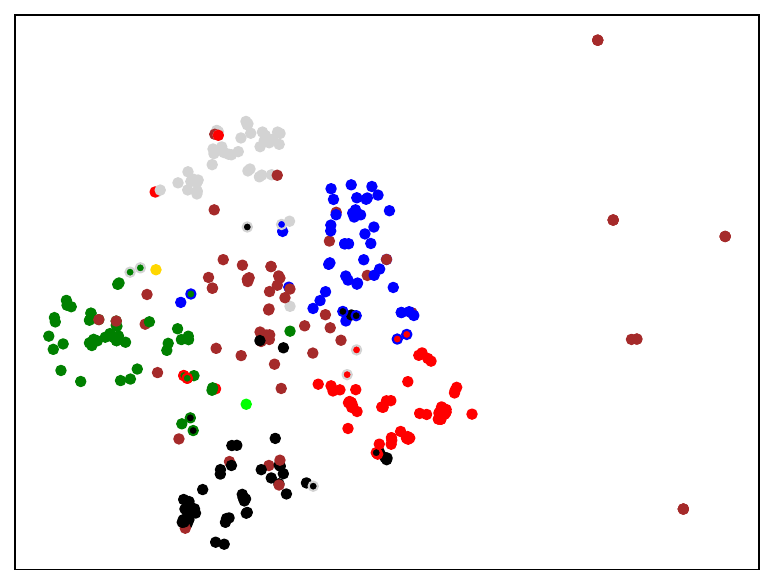} & \includegraphics[width = 5cm, height = 4cm]{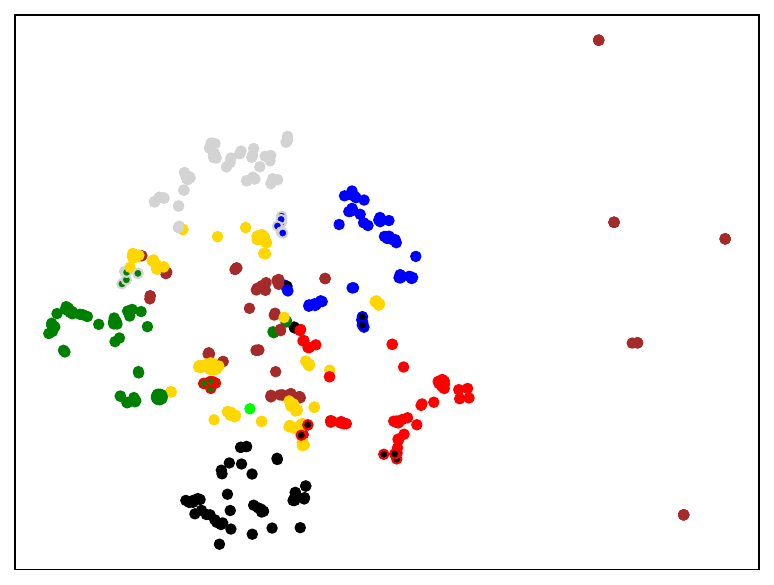}\\
(j) ONE & (k) SIR & (l) SNC\\[6pt]
\includegraphics[width = 5cm, height = 4cm]{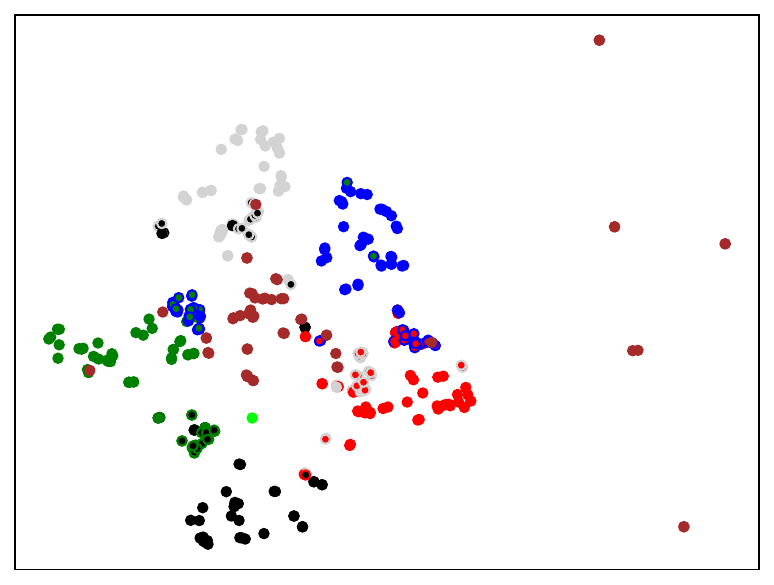} & \includegraphics[width = 5cm, height = 4cm]{images/point_clouds_shared/point_cloud_together_anch_imageSNC.pdf} & \includegraphics[width = 5cm, height = 4cm]{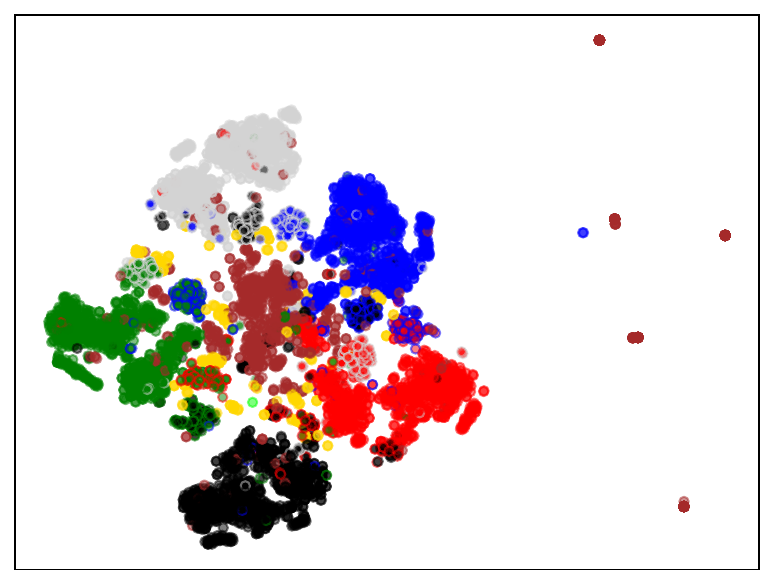}\\
(m) STX & (n) VOW & (o) combined\\[6pt]
\end{tabular}
\caption{2-Dimensional Representation of embedding space per set. This space is created with the \textit{Pre-training} model (Table~\ref{tab:accuracy_foundation}), thus cards visualised in (b) are unseen. However, a similar clustering structure emerges, clearly showing a generalised understanding of card semantics. The lime-green point denotes the embedding of an empty deck, thus each card's distance to it serves as an approximation of its absolute strength.}
\label{fig:point_clouds}
\end{figure*}

\begin{table}[htb]
\label{tab:accuracy_foundation}
\caption{Top-1 testing accuracy of models trained on a larger corpus of data. }
\resizebox{\columnwidth}{!}{
\begin{tabular}{cccc}
\toprule
& \multicolumn{3}{c}{Test data} \\
\cmidrule{2-4}
Training data & Seen sets hold-out & \textit{NEO} test-set & \textit{BRO}\\
\toprule
Pre-training & 66.49\% & 67.20\% & 55.44\% \\
+ tuning on \textit{NEO} & 58.20\% & 67.97\% & 52.25\%\\
+ tuning on \textit{BRO}& 57.97\% & 58.26\% & 62.27\%\\
\bottomrule
\end{tabular}}
\end{table}

We find that fine-tuning on \textit{NEO} barely improves accuracy on the set and does not achieve better overall results than random initialisation. This again shows that when regarding a known, fixed set of cards, generalised knowledge of card semantics seems to not affect accuracy. Additionally, when fine-tuning on one set, accuracy on the others diminishes as knowledge is forgotten. While potentially preventable by freezing layers or using small amounts of pre-training data, we omit such experiments.

\begin{figure}
    \centering
    \includegraphics[width = 0.7\columnwidth]{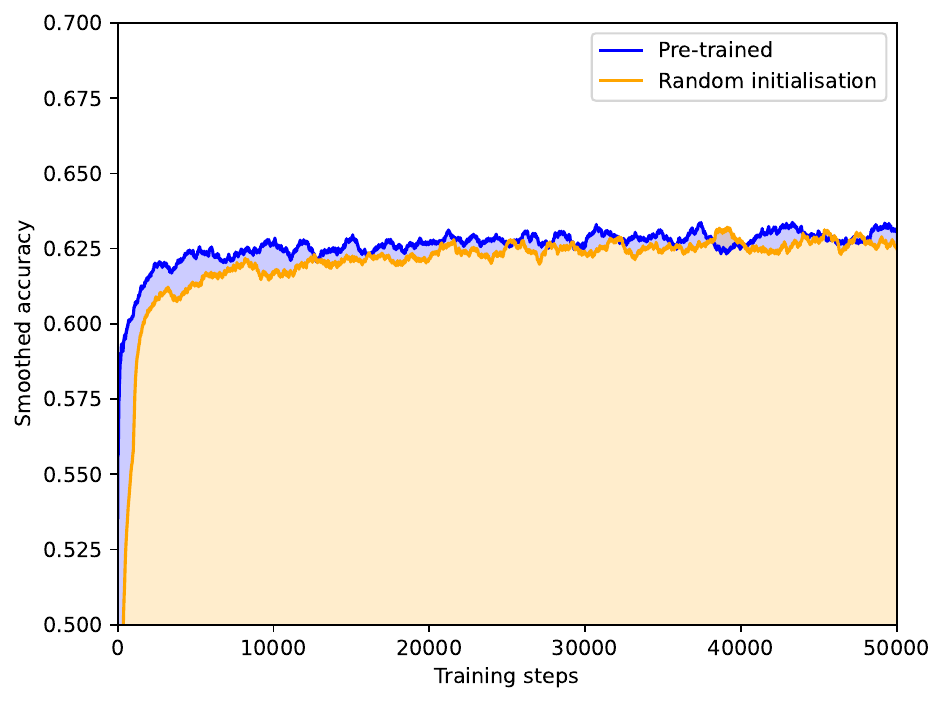}
    \caption{Accuracy curve of pre-trained model and randomly initialised model on newly seen dataset \textit{BRO}. The pre-trained model does not achieve higher peak accuracy, but meaningfully improves training speed.}
    \label{fig:tuning_bro}
\end{figure}

Fine-tuning on \textit{BRO} naturally leads to improved accuracy, as the model was not trained with these cards previously. While we do not find higher peak accuracy compared to random initialisation, using the pre-trained model leads to much quicker adaption to new data (see Figure~\ref{fig:tuning_bro}) and is thus useful for fast-paced environments.

\section{Visualisation}

Although this work mainly aims to explore card representation and general accuracy of results when using CPR to learn a universal drafting model, CPR provides the inherent advantage of an intuitively interpretative embedding space because all model predictions are directly based on minimising distances. As the embedding space is high-dimensional, we use TSNE \cite{van2008visualizing} to compute a 2-dimensional approximation (see Figure~\ref{fig:point_clouds}) and separately plot each set of cards. Each card is plotted with its respective colour in the game (see Figure~\ref{fig:magic card}), one of the most defining features. For this plot, the \textit{Pre-training} model from Section~\ref{sec:general-models} is used, but all strong models provide similar visualisations.

Highly intuitive clusters emerge in this embedding space. All colours are grouped together and multicoloured cards, visible as either two-coloured points with different borders or as golden points for more than two colours, are placed between their respective colours. All plots feature 5 points at large distance from the main plot, which are the 5 cards included in every set that provide no value to gameplay and should thus never be picked. Interestingly, the \textit{BRO} set, which consists of unseen cards for the model, exhibits similar structures as the sets which were used for training, indicating a clear semantic understanding of cards.

\section{Conclusion}

In this work, we generalised the contextual preference ranking framework \cite{bertram2021predicting} by extending it to novel card representations. The key idea is to replace the previously utilised one-hot encodings with card features, text-embeddings and auto-encoded image compressions of \textit{Magic: The Gathering} cards. Following CPR, the card representations are mapped into a high-dimensional embedding space using Siamese neural networks, in which distances model the relationship between cards and decks. We find that the input representation is largely irrelevant when modelling singular, fixed sets. In such cases, where data for all cards is available, all models with reasonable inputs perform similarly well. However, large discrepancies in accuracy emerge when aiming to understand card semantics, e.g. when predicting decisions with previously unseen cards, a common occurrence when new sets are released. Here, we find that hand-engineered features of cards, combined with meta-statistics and auto-encoded images, lead to the highest generalisation capabilities and provide quick fine-tuning opportunities when new data is available. Additionally, we find that our new representation outperforms previous research on the same dataset.

\bibliographystyle{IEEEtranS}
\bibliography{bib} 

\end{document}